\documentclass{article}

\usepackage{arxiv}
\usepackage{color,soul}
\usepackage{longtable}
\usepackage{adjustbox}
\usepackage[utf8]{inputenc} 
\usepackage[T1]{fontenc}    
\usepackage{hyperref}       
\usepackage{url}            
\usepackage{booktabs}       
\usepackage{amsmath,amssymb,amsthm, amsfonts} 
\usepackage{dsfont}
\usepackage{array,multirow}
\usepackage{nicefrac}       
\usepackage{microtype}      
\usepackage{lipsum}		
\usepackage{graphicx}
\usepackage{natbib}
\usepackage{doi}
\usepackage{algorithm}
\usepackage[noend]{algpseudocode}
\usepackage{tikz}
\usetikzlibrary{shapes, matrix, positioning}

\newtheorem{remark}{Remark}[section]

\title{Survival Multiarmed Bandits with Bootstrapping Methods}

\author{ Peter Veroutis\\
	Department of Mathematics and Statistic\\
	Concordia University\\
	Montréal, Canada \\
	\texttt{peter.veroutis@mail.concordia.ca} \\
	\And
	\href{https://orcid.org/0000-0001-5097-5269}{\includegraphics[scale=0.06]{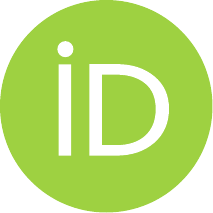}\hspace{1mm}Frédéric Godin} \\
	Department of Mathematics and Statistic\\
	Concordia Universty\\
	Montréal, Canada \\
	\texttt{frederic.godin@concordia.ca} \\
}

\renewcommand{\shorttitle}{Survival Multiarmed Bandits with Bootstrapping Methods}

\hypersetup{
pdftitle={Survival Multiarmed Bandits with Bootstrapping Methods},
pdfsubject={cs.LG},
pdfauthor={Peter Verpitos, Frédéric Godin, Chun Wang},
pdfkeywords={Risk-averse bandits, Ruin, Reinforcement Learning},
}

\begin{document}
\maketitle

\begin{abstract}
The Multiarmed Bandits (MAB) problem has been extensively studied and has seen many practical applications in a variety of fields. The Survival Multiarmed Bandits (S-MAB) open problem is an extension which constrains an agent to a budget that is directly related to observed rewards. As budget depletion leads to ruin, an agent's objective is to both maximize expected cumulative rewards and minimize the probability of ruin. This paper presents a framework that addresses such a dual goal using an objective function balanced by a ruin aversion component. Action values are estimated through a novel approach which consists of bootstrapping samples from previously observed rewards. In numerical experiments, the policies we present outperform benchmarks from the literature.

\end{abstract}

\keywords{Risk-averse bandits\and Ruin \and Reinforcement Learning}

\section{Introduction}
\label{intro}

In the Multiarmed Bandits (MAB) problem an agent learns about its environment by successively sampling from one of $K$ arms, each yielding rewards from distributions $f_1, \dots , f_K$ which are unknown to the agent \citep{sutton2018reinforcement}. Determining optimal actions requires an appropriate balance of exploration and exploitation at each stage. In the traditional setting, actions which maximize the cumulative expected reward are deemed to be optimal. The MAB framework has seen many practical applications in a wide variety of fields like healthcare, finance, machine learning and telecommunication to name a few \citep{bouneffouf2019survey}.

Recent literature has extended the bandits framework with alternative objectives such as Risk-Averse Multiarmed Bandits (RA-MAB) and Budgeted Multiarmed Bandits (B-MAB), which broaden the scope of applications of bandits models. The RA-MAB are concerned with the risk of rewards \citep{sani2012risk} and the B-MAB with a cost associated with each action that depletes a finite budget \citep{xia2017finite}. The present work addresses the Survival Multiarmed Bandits (S-MAB) framework proposed by \cite{perotto2019open}. Not to be confused with the B-MAB, the S-MAB is also constrained by a finite budget and seeks to maximize the expected cumulative reward yet while minimizing the probability of ruin. \cite{perotto2019open} outlines that the S-MAB is a multiobjective problem. The B-MAB budget is a decreasing function of stages and is independent of rewards, while in the S-MAB the budget is directly dependent on the rewards where costs take the form of negative rewards. 

Multiple applications can be represented by the S-MAB framework.
For healthcare studies, the S-MAB allows for an initial threshold $b_0$ to be set for medical trials in which the goal is to find the optimal treatment; ruin calls for the procedure to be interrupted. In the context of recommender systems, the agents budget can be used to reflect the level of user engagement, with ruin occuring if the user exits the application \citep{WANG2023102351}. If we view the initial budget in a more the literal sense of capital, a portfolio manager wishing to maximize their return from a choice of portfolios while avoid budget depletion could be viewed as an agent in the S-MAB \citep{riou2022survival}. In robotics, $b_0$ can be viewed as an initial energy level where the agent seeks to determine optimal actions while remaining alive \citep{perotto2019open}.

The current S-MAB literature has only considered bounded rewards on the interval [-1,1]. Some restrict the framework to discrete Bernoulli rewards in  \{-1,1\}, see \cite{perotto2021deciding}, \cite{perotto2021gambler},  \cite{perotto2022time} and \cite{manome2023simple}. In such works, most results rely on the Bernoulli assumption. \cite{riou2022survival} extend to continuous rewards and define the problem as the single objective to maximize expected cumulative reward until time of ruin. They offer theoretical guarantees on survival regret which rely on bounded rewards and the single objective definition. Our contribution presents a framework that can be tuned for varying degrees of ruin aversion applicable to arbitrary reward distributions.

\section{Ruin-Averse Bandits}
The S-MAB framework we consider is inspired from \cite{perotto2019open}. It includes the triplet $\{I,F,b_0\}$, where $I = \{1,\dots,K\}$ is the set of possible actions, $F = \{f_1,\dots,f_K\}$ is the set of reward distributions functions and $b_0 \in \mathbb{R}^+$ is the initial budget. 
The process evolves over discrete stages $t=1,\ldots,T$, with $T$ being the time horizon. 
At each stage $t$, an agent decides on an action $a_t \in I$. Denoting by $X_{t,k}$ the reward obtained on stage $t$ if action $k$ is selected, the agent receives reward $r_t = X_{t,a_t}$. The sequence of random vectors $\{(X_{t,1},\ldots, X_{t,K})\}^T_{t=1}$ is independent and identically distributed (i.i.d.), with $X_{t,k}$ having distribution $f_k$. All reward distributions are unknown to the agent at the onset.
The rewards impact the time-$t$ budget through $$b_t = \mathds{1}_{\{b_{t-1} > 0\}} \max (0;r_t + b_{t-1}).  $$ If the budget hits zero, ruin occurs and it remains zero afterwards.
The time until ruin is therefore defined as
\begin{equation*}
\tau = \inf\{ t > 0 : b_t = 0 \}.
\end{equation*}

Consider a ruin-averse agent whose objective is to not only maximize the expected cumulative reward, but also to avoid ruin. We assume that the agent wishes to maximize the following objective function:
\begin{equation*}
    \mathcal{O}_1(\lambda) =\mathbb{E} \left[b_T-b_0\right]- \lambda \mathbb{P}[ \tau \leq T ]
    = \mathbb{E} \left[\sum^T_{t=1} \max(r_t;-b_{t-1}) \mathds{1}_{\{\tau > t-1\}} - \lambda  \mathds{1}_ { \{\tau \leq T\}} \right].
\end{equation*}
where $\lambda$ is the hyper-parameter that drives the intensity of ruin aversion of the agent.\footnote{Using the method of Lagrange multipliers, such problem would be equivalent to solving  $\max \mathbb{E}\left[b_T-b_0\right]$ subject to \\ $\mathbb{P}[ \tau \leq T ] < c$ for some constant $c$ if arms distributions were known}.

\subsection{Action Value Definition}
The approach considered to select actions is now outlined. From the law of total expectation
\begin{align*}
    \mathcal{O}_1(\lambda) =&  \mathbb{E} \bigg[ \sum^{t-1}_{s=1} \max(r_s;-b_{s-1}) \mathds{1}_{\{ \tau > s-1\}}  -\lambda \mathds{1}_{ \{ \tau \leq t-1 \} } 
    \\ &+ \mathds{1}_{ \{\tau > t-1 \} }\underbrace{\mathbb{E}\left[ \sum^T_{s=t} \max(r_s;-b_{s-1}) \mathds{1}_{\{ \tau > s-1\}} - \lambda\mathds{1}_ { \{\tau \leq T\}} \bigg\vert a_{1:(t-1)}, r_{1:(t-1)} \right]}_{=: \mathcal{O}_t(\lambda)}\bigg].
\end{align*}

As such the greedy action is that which is estimated to maximize $\mathcal{O}_t(\lambda)$ if selected repeatedly until $T$. Denote the time-$t$ action value of arm $k$ as
\begin{equation}
\label{eq:timekvaluefunct}
   \mathcal{O}^{(k)}_t(\lambda) = \mathbb{E}\left[ \sum^T_{s=t} \max(r_s,-b_{s-1})\mathds{1}_{\{ \tau > s-1\}} - \lambda\mathds{1}_{ \{t \leq \tau \leq T\}} \bigg\vert a_{t:T}=k, b_{t-1} \right].
\end{equation}
Note that the quantity inside the expectation in \eqref{eq:timekvaluefunct} is
\begin{equation*}
\sum^T_{s=t} \max(r_s,-b_{s-1})\mathds{1}_{\{ \tau > s-1\}} - \lambda\mathds{1}_{ \{t \leq \tau \leq T\}} =
    \begin{cases}
        \sum^T_{s=t} r_s \text{ if } \tau > T,
         \\ -b_{t-1}-\lambda \text{ otherwise.}
    \end{cases}
\end{equation*}
On any stage $t$, the greedy action is defined as $\underset{k=1,\ldots,K}{\arg\max} \text{ }\hat{\mathcal{O}}^{(k)}_t(\lambda)$, with $\hat{\mathcal{O}}^{(k)}_t(\lambda)$ being an estimate of ${\mathcal{O}}^{(k)}_t(\lambda)$.
\subsection{Bootstrapping for Action Value Estimation}
The approach we consider to estimate quantities $\hat{\mathcal{O}}^{(k)}_t(\lambda)$ is bootstrapping.
Assuming arm $k$ has been pulled at least once by time $t-1$ and that no ruin has yet occurred, the non-empty set of observed rewards for that arm is denoted
\begin{equation*}
\mathcal{Z}_{t}^{k} = \{r_{i} \:|\: a_i = k, \: 1 \leq i \leq t-1 \}.
\end{equation*}
To compute the time-$t$ bootstrapping  estimate $\hat{\mathcal{O}}^{(k)}_t(\lambda)$, we simulate $M$ paths through sampling with replacement from set $\mathcal{Z}_{t}^{k}$. The $m^{th}$ path is denoted as the sequence $\tilde{r}^{(m)}_{t:T} = \{\tilde{r}^{(m)}_t,\ldots,\tilde{r}^{(m)}_T$\}, and the associated time of ruin is denoted $\tilde{\tau}^{(m)}_t = \inf\{ s \geq t : b_s = 0 \}$, $m=1\ldots,M$.
From such simulated paths, sample averaging is considered
\begin{equation} \label{eq:4}
\hat{\mathcal{O}}^{(k)}_t(\lambda) = \frac{1}{M} \sum^M_{m=1} \left[\mathds{1}_{ \{  \tilde{\tau}^{(m)}_t \leq T\}}(-b_{t-1}-\lambda)  + \mathds{1}_{ \{  \tilde{\tau}^{(m)}_t > T\}}\sum^T_{s=t} \tilde{r}^{(m)}_s \right].
\end{equation}
\begin{remark}
    Equation \eqref{eq:4} entails bootstrapping from a discrete distribution. Smooth bootstrapping (e.g. bootstrapping from kernel densities) was also tested. However, as in \cite{de1992smoothing} it proved more computationally expensive, did not lead to a material improvement in the estimators, and requires to consider an additional smoothing parameter that needs to be tuned.
\end{remark}
\subsection{An Upper Confidence Bound Algorithm}
Algorithms we propose to select respective actions is inspired from the Upper Confidence Bound (UCB) approach \citep[see][]{auer2002finite}. Such scheme tackles the exploration-exploitation dilemma by incorporating a term to the objective function that encourages the selection of actions that have not been sampled extensively and for which statistical confidence about their value is low. This promotes the idea of \textit{optimism in the face of uncertainty}, which entail exploring actions based not only on their past performance, but also on their potential to reveal themselves as optimal once statistical uncertainty is resolved.

Denoting by $N_{t}(k)$ the number of times action $k$ has been sampled by time $t-1$, the UCB term we consider in a first algorithm has the form $\sqrt{\frac{\alpha\ln{(t)}}{N_{t}(k) }}$ for some $\alpha >0$, as in \cite{auer2002finite}. Such term is a decreasing function of $N_{t}(k)$, encouraging the selection of actions that were sampled less extensively. Furthermore, for a fixed $N_{t}(k)$, such term tends to infinity as $t$ goes to infinity, ensuring that all actions are sampled infinitely often would we have $T\rightarrow\infty$. The
the \texttt{UCB} algorithm of \cite{auer2002finite} entails selecting actions according to
\begin{equation*}\label{eq:UBCPolicy}
a_t = \underset{k=1,\ldots,K}{\arg\max} \left[  Q_t^{(k)} + \sqrt{\frac{\alpha\ln{(t)}}{N_{t}(k)}} \right]
\end{equation*}
where $Q_t^{(k)} = \frac{\sum^{t-1}_{s=1} r_s \mathds{1}_{ \{ a_s = k\} } }{ \sum^{t-1}_{s=1} \mathds{1}_{ \{ a_s = k\} } }$ is the traditional estimate of the value for action $k$. However, to account for the risk of ruin, we propose instead the \texttt{$\lambda$-RuinAverse} policy which selects actions according to 
\begin{equation*}\label{eq:UBCtRuinAverse}
a_t = \underset{k=1,\ldots,K}{\arg\max} \left[  \mathcal{O}^{(k)}_t(\lambda) + \sqrt{\frac{\alpha\ln{(t)}}{N_{t}(k)}} \right],
\end{equation*}
i.e. by replacing $Q_t^{(k)}$ with $\mathcal{O}^{(k)}_t(\lambda)$. Such policy is presented in the Algorithm \ref{algo:auto_thresh} box.
\begin{algorithm}[H]
    \caption{ $\lambda$-RuinAverse policy }
    \label{algo:auto_thresh}
    \begin{algorithmic} 
        \State \textbf{Inputs:} \( \lambda \geq 0, \alpha > 0 \).
        \For{$t = 1,\dots,T$}
            \For{$k = 1,\dots,K$}
                \For{$m = 1,\dots,M$}
                \State Sample a path $\tilde{r}^{(m)}_{t:T}$ of $T-t+1$ rewards with replacement from $\mathcal{Z}_{t}^{k}$               
                \EndFor 
                \State Compute $\hat{\mathcal{O}}^{(k)}_t(\lambda)$ based on \eqref{eq:4}
            \EndFor           
            \State $a_t = \underset{k=1,\ldots,K}{\arg\max} \left[  \hat{\mathcal{O}}^{(k)}_t(\lambda) + \sqrt{\frac{\alpha\ln{(t)}}{N_{t}(k)}} \right]$
            \State Observe $r_t$ from distribution $f_{a_t}$
            \State $b_{t} \leftarrow  r_t + b_{t-1}$
            \If{$b_{t} \leq 0$}
                \State $b_t,\ldots,b_T \leftarrow 0$
                \State Break loop
            \EndIf
        \EndFor
    \end{algorithmic}
\end{algorithm}

\cite{perotto2022time} proposes an alternate UCB-inspired-term where the agent favors exploitation when its budget is low, and exploration when its budget is high. By replacing $t$ with $b_t + 1$ in the UCB term, optimal actions in such strategy, which we refer to as \texttt{UCBBudget}, are chosen according to\footnote{Originally used for discrete Bernoulli arms, the UCB term in  \cite{perotto2022time}  is $\sqrt{\frac{\alpha\ln{(b_t)}}{N_{t}(a)}}$. Yet when extending to continuous arms, $b_t$ has the possibility of being less than 1 while the agent is still alive. To ensure that the argument of the logarithm is positive, we replace $b_t$ with $b_t$ + 1.}
\begin{equation*}
a_t = \underset{k=1,\ldots,K}{\arg\max} \left[ Q_{t}^{(k)} + \sqrt{\frac{\alpha\ln{(b_t + 1)}}{N_{t}(k)}} \right].
\end{equation*}

We propose a second policy analogous to the above, which is referred to as \texttt{$\lambda$-RuinAverseUCBBudget}, and which again replaces $Q_t^{(k)}$ with $\mathcal{O}^{(k)}_t(\lambda)$, while still using a budget-aware UCB-inspired term. It selects optimal actions through 
\begin{equation*}\label{eq:10}
a_t = \underset{k=1,\ldots,K}{\arg\max} \left[  \mathcal{O}^{(k)}_t(\lambda) + \sqrt{\frac{\alpha\ln{(b_t + 1)}}{N_{t}(k)}} \right].
\end{equation*}

\begin{remark}
Epsilon-greedy-inspired policies were also assessed in unreported experiments. However, due to insufficient exploration in early stages (i.e. when $t$ is small), these policies failed to provide good estimates of ${O}_t^{(k)}(\lambda)$ for such early stages during which decision making is crucial for the survival of the agent. As such, the epsilon-greedy approach was not pursued further.
\end{remark}

\section{Numerical Experiments}
Numerical experiments are conducted by performing Monte Carlo simulations in an eight-arm environment.
The classic \texttt{UCB} policy and the \texttt{UCBBudget} policy originally proposed by \cite{perotto2022time} are compared to our proposed algorithms, namely the \texttt{$\lambda$-RuinAverse} and the \texttt{$\lambda$-RuinAverseUCBBudget}.
\subsection{Setup}
The agent is presented with an environment of eight normally distributed arms in which 1000 runs are repeated per policy, with a time horizon of $T=500$ stages. 
On each stage, to perform bootstrapping estimations, $M = 100$ paths are generated. The UCB hyper-parameter is set to $\alpha = 10$.
At the start of each run, the budget is initialized to $b_0 = 0.5$, and mean and variance parameters for rewards are generated randomly from uniform and gamma\footnote{Gamma($\alpha$, $\beta$) denotes the gamma distribution with pdf $f(x) = \frac{\beta^\alpha}{\Gamma(\alpha)} x^{\alpha-1} e^{-\beta x},
\quad x > 0$.} distributions as follows

\begin{table}[ht]
  \caption{Eight normal arms environment.}
  \label{example_table}
  \centering
  \begin{tabular}{ccc}
    \toprule
    Rewards & Mean & Variance  \\
    \midrule
    $r_t \sim \mathcal{N}(\mu_k, \sigma_k^2)$ & $\mu_k \sim \text{Unif}(-0.01, 0.01)$ & $\sigma_k \sim \text{Gamma}(1, 10)$ \\
    \bottomrule
  \end{tabular}
\end{table}

Distributions for the simulated mean and variance parameters were chosen to create an environment where the budget never strays too far away from zero and the risk of ruin always remain material.


\subsection{Performance}
\begin{figure}[H]
\caption{Performance over stages}
    \centering
    \includegraphics[width=\textwidth]{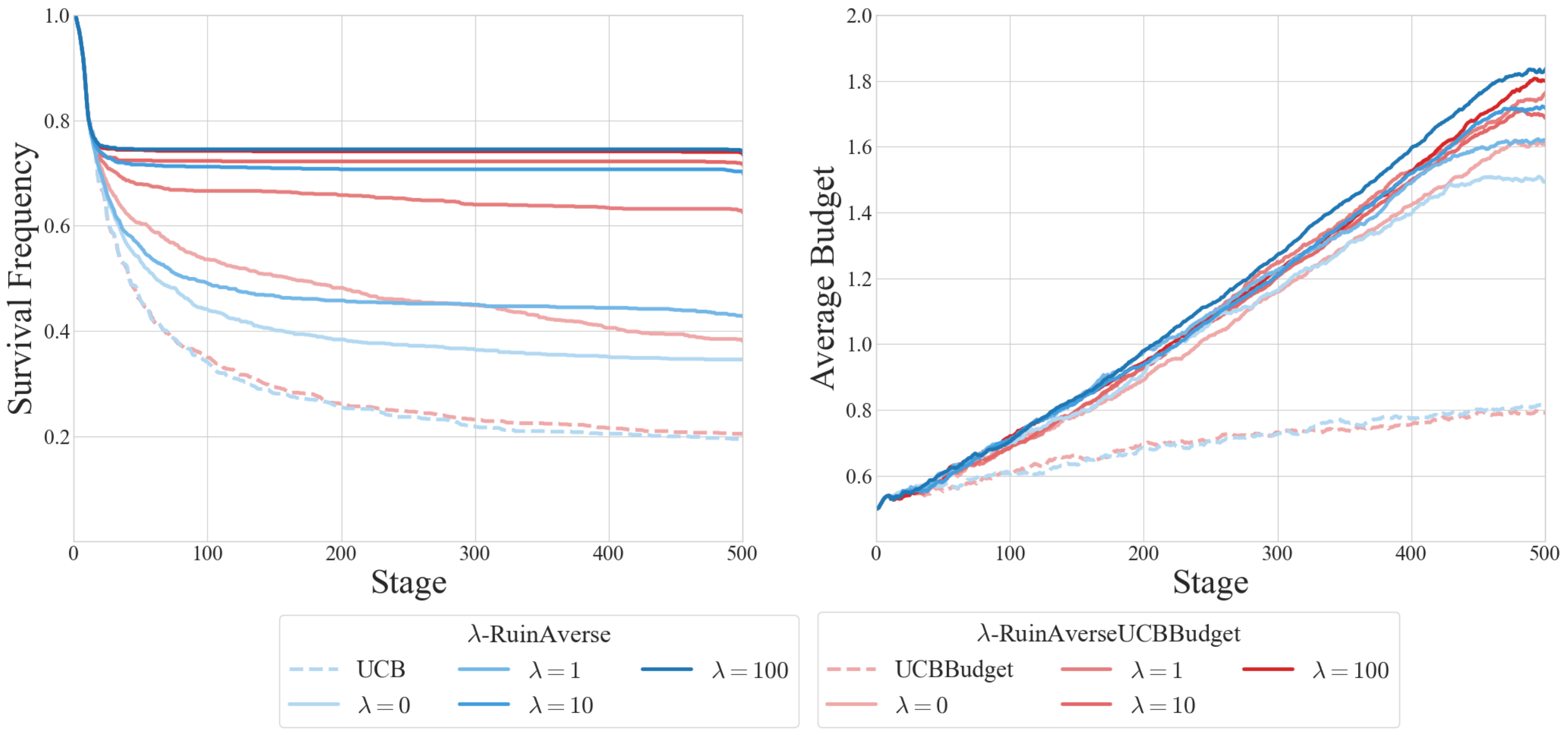}
    \label{fig:metrics}
\end{figure}

The curves for the survival frequency and the average budget over runs versus each stage $t$ are graphed in Figure \ref{fig:metrics}. For the various stages $t$, survival curves present the proportion of non-ruined runs by stage $t$, while average budget curves present the mean budget across runs at stage $t$. Different values of $\lambda$ are tested for the \texttt{$\lambda$-RuinAverse} policy (blue curves) and the \texttt{$\lambda$-RuinAverseUCBBudget} policy (red curves). As curves get progressively darker in color the agent becomes increasingly ruin averse. 
\newpage

In this environment, the budget-aware benchmark \texttt{UCBBudget} has similar performance to the classic \texttt{UCB}. Moreover, for all $\lambda$ values considered, both \texttt{$\lambda$-RuinAverse} and \texttt{$\lambda$-RuinAverseUCBBudget} policies that are ruin-aware significantly outperform their respective non-ruin-aware counterparts. 
As expected, factoring in estimated costs of ruin in action selection improves performance.
With an increase in $\lambda$, survival curves decrease less rapidly and flatten at early stages. This highlights
high avoidance of risky actions as soon as they are identified by policies with high ruin aversion.

In Figure \ref{fig:metrics}, the average budget appear to flatten in the last $50$ stages. This is because the cost of ruin becomes large relative to the few remaining awards that can be collected by the agent when nearing the time horizon $T$, and thus the ruin-aware agent decides to behave safely in the final stages. Indeed, such trend is not observed for any of the two benchmark that do not factor in ruin aversion and do not need to weight the cost of ruin against expected future rewards. Applying a discount factor on the cost of ruin would dampen this phenomenon as late-stage ruin instances would be less costly.

\begin{table}[ht]
  \caption{Performance metrics at stage $T = 500$}
  \label{Tmetrics}
  \centering
  \begin{tabular}{llcccc}
    \toprule
    && \multicolumn{2}{c}{$\lambda$-RuinAverse} & \multicolumn{2}{c}{$\lambda$-RuinAverseUCBBudget}  \\
    \cmidrule{3-6}
    \multicolumn{2}{l}{Policy} & Survival Frequency & Average Budget & Survival Frequency & Average Budget \\
    \midrule
    \multicolumn{2}{l}{UCB}& 0.195 & 0.809 & - & - \\
    \multicolumn{2}{l}{UCBBudget}& - & - & 0.204 & 0.792 \\
    \midrule
    \multirow{5}{*}{$\lambda := $\(\left\{\begin{array}{l} 0 \\ 1 \\ 10 \\ 10^2 \\ 10^3 \end{array}\right.\)} && 0.346 & 1.492 & 0.381 & 1.607 \\
    && 0.427 & 1.621 & 0.624 & 1.766 \\
    && 0.697 & 1.715 & 0.713 & 1.685 \\
    && 0.736 & 1.838 & 0.733 & 1.800 \\
    && 0.746 & 1.742 & 0.743 & 1.739 \\
    \bottomrule
  \end{tabular}
\end{table}















Table \ref{Tmetrics} lists the performance metrics at time horizon $T=500$ for all tested policies. For smaller $\lambda$, the \texttt{$\lambda$-RuinAverseUCBBudget} policy outperforms \texttt{$\lambda$-RuinAverse} in both survival frequency and average budget metrics.
As $\lambda$ grows increasingly large, the survival frequency at stage $T = 500$ stabilizes, with the policy converging to a purely ruin-averse one.

Moreover, we observe that results for some values of $\lambda$ dominate
that of others, with smaller values $\lambda$ = 0, 1, 10 all exhibiting both lower survival frequencies and
average budgets than $\lambda = 10^2$. This occurs because the impact of ruin is twofold; it is not only directly penalized, but it also prevents the agent from collecting any additional reward after the time of ruin. Hence, some minimal degree of prudence is warranted as it allows both avoiding the cost of ruin and collecting more rewards, thereby increasing the expected budget. However, for larger values of $\lambda$, no dominance relationship is established, with a trade-off between the survival frequency and the expected cumulative rewards being observed, e.g. when comparing $\lambda=10^2$ to $\lambda=10^3$. As mentioned in \cite{perotto2019open}, optimizing simultaneously for the final expected budget and the survival frequency produces a Pareto-optimal frontier, and $\lambda$ can be tuned depending the extent to which both objectives are prioritized relative to each other.

\section{Conclusion and Future Work}
This work presents a framework to tackle the Survival Multi-Armed Bandits (S-MAB) problem. A formulation of the problem with an objective function embedding both expected rewards and a ruin cost component is proposed, which entails that the agent is both ruin-averse and reward-seeking. The framework can handle general reward distributions and is not limited to bounded or discrete distributions as in other recent literature works.
In our proposed approach, a UCB term manages the exploitation-exploration dilemma by linking the appetite for exploration to either (i) uncertainty in the action value estimates, or (ii) the current budget.
Action value estimates, which appear in the action selection formulas, are estimated with a novel approach consisting of generating bootstrapped samples made of previously observed rewards.  Simulation experiments highlight that our proposed algorithms lead to a significant improvement in performance over literature benchmarks in terms of survival frequency and the expected cumulative reward. Results reveal that strategies with insufficient ruin aversion are often dominated by these exhibiting more prudent behavior. Nevertheless, for sufficiently large ruin aversion $\lambda$, a trade-off is observed between the expected cumulative reward and the probability of ruin. 

Future research may explore how to perform online tuning of the ruin aversion parameter $\lambda$ over the course of a run. Ideas from the learned optimizers literature could be leveraged to a propose a policy in which $\lambda$ is learned rather than user-specified, see for example \cite{lan2024learningoptimizereinforcementlearning}. Another direction for further work would be to meaningfully define regret for the S-MAB problem, which is non-trivial due to multiple objectives being simultaneously pursued by the agent. Properties of regret for the proposed action selection algorithms could then be studied. 
This would also allow examining if the considered UCB action selection formula driving the exploration can be refined in our setting.

\section*{Acknowledgements}
Financial support from Natural Sciences and Engineering Research Council of Canada (Veroutis: USRA award, Godin: RGPIN-2024-04593) is gratefully acknowledged.

\bibliographystyle{unsrtnat}
\bibliography{references}

\end{document}